\documentclass[letterpaper]{article}
\usepackage{spconf,amsmath,graphicx}
\usepackage{amsfonts,amssymb}
\usepackage{algorithm}
\usepackage[noend]{algpseudocode}
\usepackage[colorlinks=true, citecolor={blue}, bookmarks=false, breaklinks]{hyperref}
\usepackage{url}

\usepackage{breakurl}
\usepackage{multirow}
\usepackage{adjustbox}
\usepackage[numbers]{natbib}

\title{CAPSULE-FORENSICS: USING CAPSULE NETWORKS TO DETECT FORGED IMAGES AND VIDEOS}

\name{Huy H. Nguyen$^{\star}$, Junichi Yamagishi$^{\star\dagger\ddagger}$, and Isao Echizen$^{\star\dagger}$}
\address{$^{\star}$SOKENDAI (The Graduate University for Advanced Studies), Kanagawa, Japan\\
$^{\dagger}$National Institute of Informatics, Tokyo, Japan\\
$^{\ddagger}$The University of Edinburgh, Edinburgh, UK\\
\small{Email: \{nhhuy, jyamagishi, iechizen\}@nii.ac.jp}}

\begin{document}
\maketitle
\begin{abstract}
Recent advances in media generation techniques have made it easier for attackers to create forged images and videos. State-of-the-art methods enable the real-time creation of a forged version of a single video obtained from a social network. Although numerous methods have been developed for detecting forged images and videos, they are generally targeted at certain domains and quickly become obsolete as new kinds of attacks appear. The method introduced in this paper uses a capsule network to detect various kinds of spoofs, from replay attacks using printed images or recorded videos to computer-generated videos using deep convolutional neural networks. It extends the application of capsule networks beyond their original intention to the solving of inverse graphics problems.
\end{abstract}


\section{INTRODUCTION}
\label{sec:intro}
Forged images and videos can be used to bypass facial authentication and to create fake news media. The quality of manipulated images and videos has seen significant improvement with the development of advanced network architectures and the use of large amounts of training data. This has dramatically simplified the creation of facial forgeries. Nowadays, the only thing needed to create a forged facial image is simply a short video of the target person~\cite{thies2016face2face, kim2018deep} or an ID photo~\cite{averbuch2017bringing, chung2017you}. The techniques developed by Chung et al.~\cite{chung2017you} and Suwajanakorn et al.~\cite{suwajanakorn2017synthesizing} can improve the ability of attackers to learn the mapping between speech and lip motion, enabling the creation of fully synthesized audio-video data for any person. In this age of social networks serving as major sources of information, fake news with manipulated multimedia can quickly spread and have significant effects. The “deepfake” phenomenon~\cite{deepfake} is a good example of this threat\textemdash any person with a personal computer can create videos incorporating the facial image of any celebrity by using a human image synthesis technique based on artificial intelligence.

Several countermeasures have been proposed to deal with manipulated images and videos. However, most of them are aimed at particular types of attacks. For example, local binary pattern (LBP)-based methods~\cite{chingovska2012effectiveness, de2013can} are effective against replay attacks in which the attacker places a printed photo or displays a video on a screen in front of the camera. However, the eyes-focused method designed to detect a deepfake forgery~\cite{li2018ictu} can fail with the replay attack when the video displayed is of the actual target person. Other methods have more generalized ability; for instance, Fridrich and Kodovsky's method~\cite{fridrich2012rich} can be applied for both steganalysis and detecting facial reenactment videos. However, its performance on secondary tasks is limited in comparison with task-specific methods like that of Rossler et al.~\cite{rossler2018faceforensics}. Moreover, while some methods can detect a single forged image~\cite{rossler2018faceforensics, nguyen2018modular, afchar2018mesonet}, others require video input~\cite{li2018ictu}.

This paper presents a method that uses a capsule network to detect forged images and videos in a wide range of forgery scenarios, including replay attack detection and (both fully and partially) computer-generated image/video detection. This is pioneering work in the use of capsule networks~\cite{hinton2011transforming, sabour2017dynamic, hinton2018matrix}, which were originally designed for computer vision problems, to solve digital forensics problems. A comprehensive survey of state-of-the-art related work and intensive comparisons using four major datasets demonstrated the superior performance of the proposed method.

\section{RELATED WORK}
\label{sec:related}
In this section, we group forgery detection approaches into replay attack detection and computer-generated image/video detection on the basis of the features used and their target. Note that some approaches are two-fold while others are applicable only to certain types of attacks. We also provide some basic information about capsule networks and the dynamic routing algorithm that made this kind of network practical.

\subsection{Replay Attack Detection}
Prior to the current deep learning era, LBP methods were the primary defense against replay attacks~\cite{chingovska2012effectiveness, de2013can}. The method introduced by Kim et al.~\cite{kim2015face}, which is based on local patterns of the diffusion speed (“local speed patterns”), achieves higher accuracy than that of LBP-based methods. Now, with the introduction of deep learning, the ability to detect replay attacks has been greatly improved. The method of Yang et al.~\cite{yang2014learn} uses a support vector machine to classify features extracted by a pre-trained convolutional neural network (CNN). That of Menotti et al.~\cite{menotti2015deep} uses a similar procedure but optimizes the filters in an available high-performance CNN architecture. The method of Alotaibi and Mahmood~\cite{alotaibi2017deep} uses nonlinear diffusion based on an additive operator splitting scheme in their own CNN. The recently introduced method of Ito et al.~\cite{ito2017recent} leverages a pre-trained CNN and utilizes the whole image instead of only the extracted face region.

\subsection{Computer-Generated Image/Video Detection}
There are several state-of-the-art methods for detecting images or videos generated by computer using, for example, a deepfake technique for face swapping~\cite{deepfake}, the Face2Face method for facial reenactment~\cite{thies2016face2face}, or the deep video portraits technique~\cite{kim2018deep} for the purpose of forgery. Fridrich and Kodovsky~\cite{fridrich2012rich} proposed a hand-crafted-feature noise-based approach for steganalysis that can also be used for forgery detection. Cozzolino et al.~\cite{cozzolino2017recasting} implemented a CNN version of this approach. Raghavendra et al.~\cite{raghavendra2017transferable} described the special case of fine-tuning two available CNNs while Rossler et al.~\cite{rossler2018faceforensics} used only one CNN. Bayar and Stamm~\cite{bayar2016deep}, Rahmouni et al.~\cite{rahmouni2017distinguishing}, Afchar et al.~\cite{afchar2018mesonet}, Quan et al.~\cite{quan2018distinguishing}, and Li et al.~\cite{li2018ictu} proposed their own networks. Li et al.'s network~\cite{li2018ictu}, for example, is video based and uses temporal information to detect eye blinking. We used a hybrid approach~\cite{nguyen2018modular} incorporating part of a pre-trained VGG (Visual Geometry Group)-19 network~\cite{simonyan2014very} and a proposed CNN. Zhou et al.~\cite{zhou2017two} proposed a two-stream network.

\subsection{Capsule Networks}
Hinton et al.~\cite{hinton2011transforming} addressed the limitations of CNNs applied to inverse graphics tasks and laid the foundation for a more robust ``capsule'' architecture in 2011. However, this complex architecture could not be effectively implemented at the time due to the lack of an efficient algorithm and the limitations of computer hardware. Instead, easy-to-design easy-to-train CNNs became widely used. Now, with the introduction of the dynamic routing algorithm~\cite{sabour2017dynamic} and the expectation-maximization routing algorithm~\cite{hinton2018matrix}, capsule networks have been implemented with remarkable initial results. Two recent studies demonstrated that, with the agreement between capsules calculated by the dynamic routing algorithm, the hierarchical pose relationships between object parts can be well described. This has improved the accuracy of vision tasks.
Application of a capsule network to the forensics task, the focus of this paper, is a challenging problem. However, the agreement between capsules achieved by using the dynamic routing algorithm could boost detection performance on complex and nearly flawless forged images and videos.

\section{CAPSULE-FORENSICS}
\label{sec:capsule}
\subsection{Overview}
\begin{figure}[th!]
\begin{center}
\includegraphics[width=86mm]{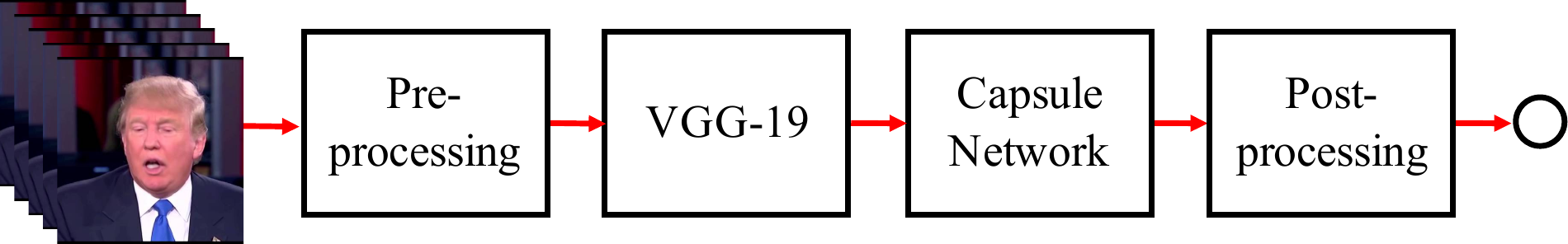}
\caption{Overview of proposed method.}
\label{figure:overview}
\end{center}
\end{figure}

The proposed method (Fig.~\ref{figure:overview}) works for both images and videos. For video input, the video is split into frames in the pre-processing phase. The classification results (posterior probabilities) are then acquired from the frames. The probabilities are averaged in the post-processing phase to get the final result. The remaining parts are identical to the input image.

In the pre-processing phase, faces are detected and scaled to $128 \times 128$. Like we did in our previous work~\cite{nguyen2018modular}, we use part of the VGG-19 network~\cite{simonyan2014very} to extract the latent features, which are the inputs to the capsule network. Unlike we did in our previous work, we take the output of the third maxpooling layer instead of three outputs before the ReLU layers. We do this because we need to reduce the size of the inputs to the capsule network. 

\subsection{Capsule Design}
\begin{figure}[th!]
\begin{center}
\includegraphics[width=78mm]{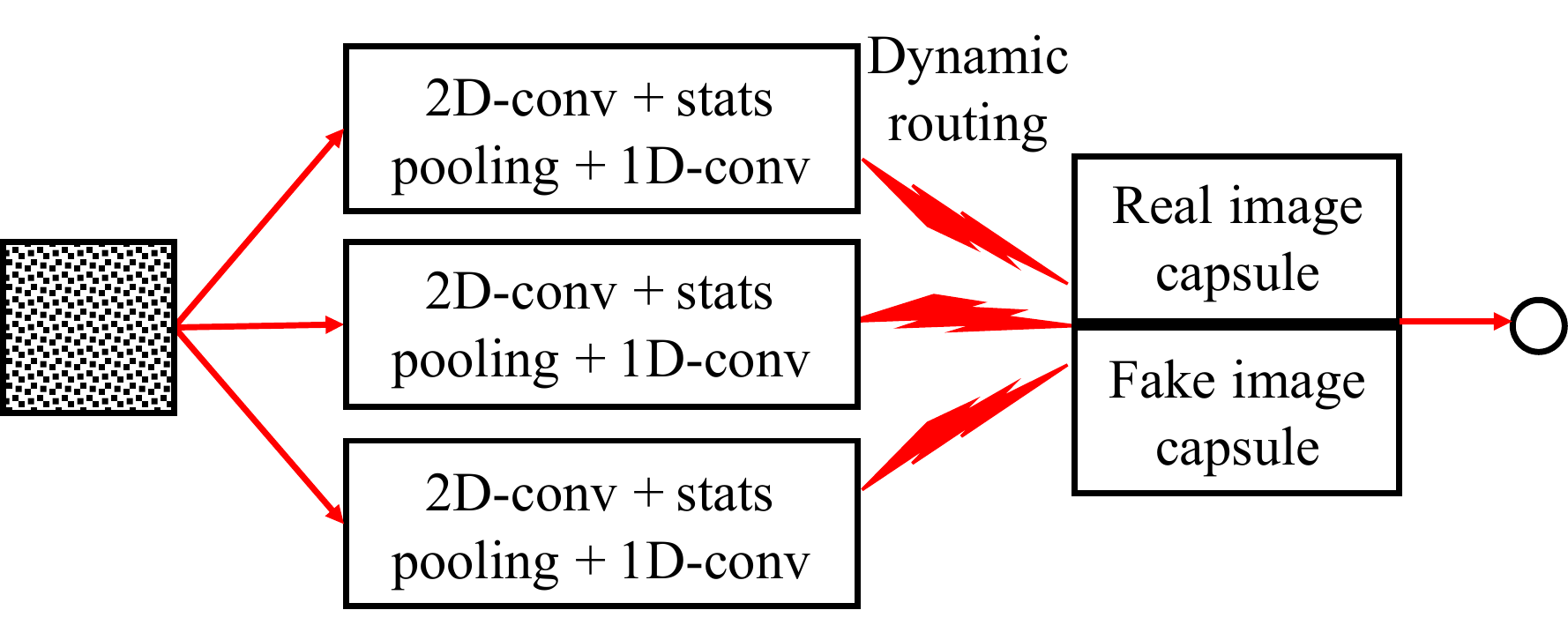}
\caption{Overall design of capsule-forensics network.}
\label{figure:capsule}
\end{center}
\end{figure}

\begin{figure}[th!]
\begin{center}
\includegraphics[width=70mm]{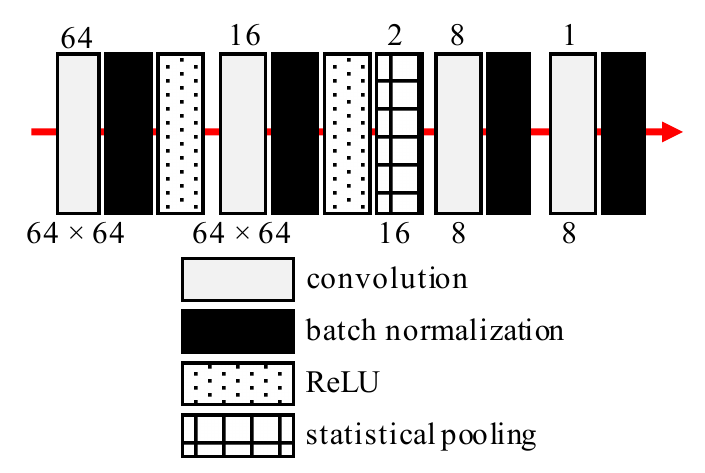}
\caption{Detailed design of primary capsule. Upper numbers indicate number of filters (depth) while lower number indicate size of outputs of corresponding filters.}
\label{figure:cnn}
\end{center}
\end{figure}

\begin{algorithm}
\caption{Dynamic routing between capsules.}
\label{alg:routing}
\begin{algorithmic}
\Procedure{Routing}{$\textbf{u}_{j|i}, W, r$}
    \State $\hat{W}\gets W + rand(size(W))$
    \State $\hat{\textbf{u}}_{j|i}\gets \hat{W}_i squash(\textbf{u}_{j|i})$ \Comment{$W_i \in R^{m \times n}$}
    \For {all input capsule $i$ and all output capsules $j$}
        \State $b_{ij}\gets 0$
    \EndFor
    \For {$r$ iterations}
        \State \textbf{for} all input capsules $i$ \textbf{do} $c_i\gets softmax(b_i)$
        \State \textbf{for} all output capsules $j$ \textbf{do} $\textbf{s}_j\gets \sum_{i} c_{ij}\hat{\textbf{u}}_{j|i}$
        \State \textbf{for} all output capsules $j$ \textbf{do} $\textbf{v}_j\gets squash(\textbf{s}_j)$
        \For {all input capsules $i$ and output capsules $j$}
            \State $b_{ij}\gets b_{ij} + \hat{\textbf{u}}_{j|i}\cdot \textbf{v}_j$
        \EndFor
    \EndFor
    \State \textbf{return} $\textbf{v}_j$
\EndProcedure
\end{algorithmic}
\end{algorithm}

\begin{figure}[ht!]
\begin{center}
\includegraphics[width=82mm]{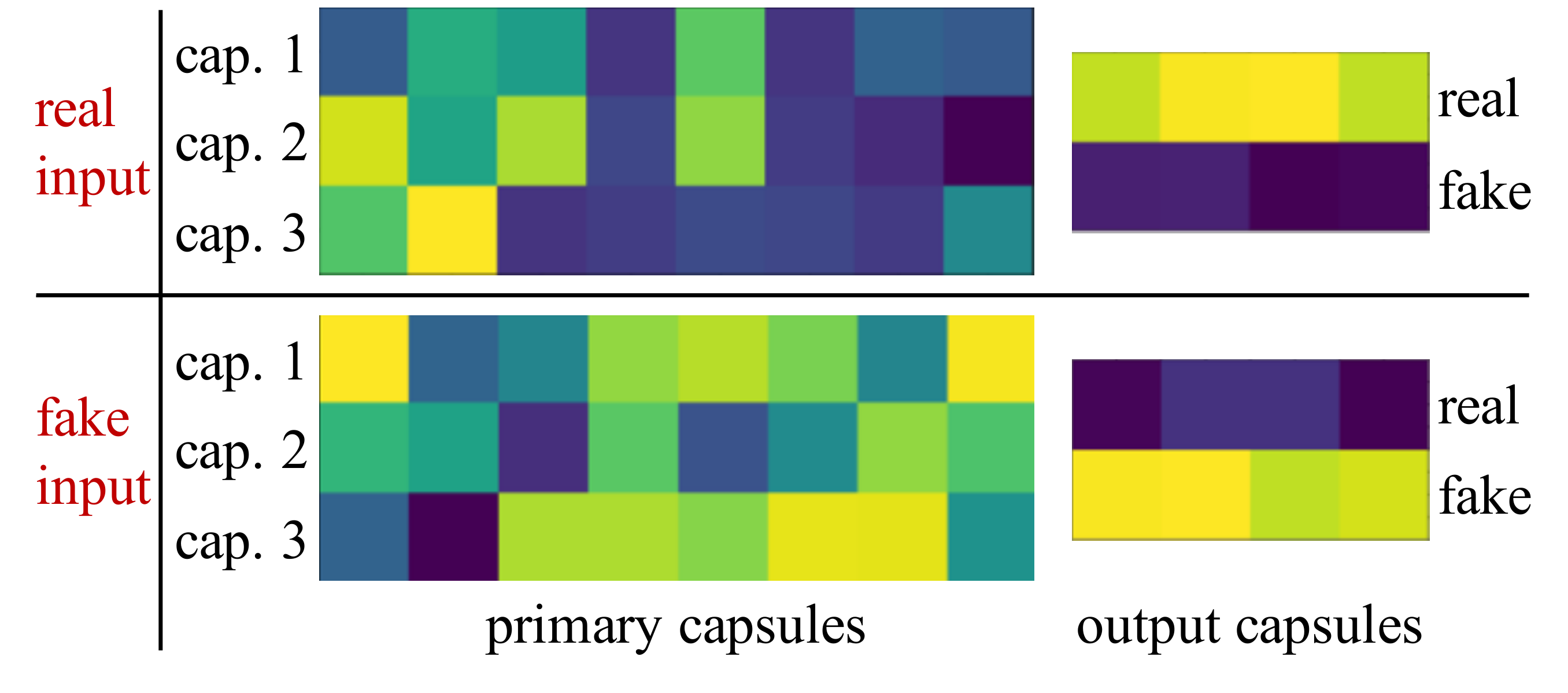}
\caption{Average results calculated by primary capsules and output capsules from real and fake images generated with Face2Face method~\cite{thies2016face2face}. Three primary capsules have significantly different reactions between real and fake inputs. Although their weights are also different, there is strong agreement in the output capsules.}
\label{figure:visualization}
\end{center}
\end{figure}

The proposed network consists of three primary capsules and two output capsules, one for real and one for fake images (Fig.~\ref{figure:capsule}). The latent features extracted by part of the VGG-19 network~\cite{simonyan2014very} are the inputs, which are distributed to the three primary capsules (Fig.~\ref{figure:cnn}). As in our previous work~\cite{nguyen2018modular}, statistical pooling, which is important for forgery detection, is used. The outputs of the three capsules ($\textbf{u}_{j|i}$) are dynamically routed to the output capsules ($\textbf{v}_j$) for $r$ iterations using Algorithm~\ref{alg:routing}. The network has approximate 2.8 million parameters, a relatively small number for such networks. We slightly improved the algorithm of Sabour et al.~\cite{sabour2017dynamic} by adding Gaussian random noise to the 3-D weight tensor $W$ and applying one additional $squash$ (equation~\ref{eq:squash}) before routing by iterating. The added noise helps reduce over-fitting while the additional equation keeps the network more stable. The outputs of the primary and output capsules are illustrated in Fig.~\ref{figure:visualization}.

\begin{equation}
\label{eq:squash}
 \textbf{v}_j = squash(\textbf{s}_j) = \frac{\|\textbf{s}_j\|^2}{1 + \|\textbf{s}_j\|^2}\frac{\textbf{s}_j}{\|\textbf{s}_j\|}
\end{equation}

Unlike Sabour et al.'s work~\cite{sabour2017dynamic}, we use the cross-entropy loss function:

\begin{equation}
\label{eq:loss}
 L = -\left(y\log(\hat{y}) + (1 - y)\log(1 - \hat{y}) \right),
\end{equation}
where $y$ is the ground truth label and $\hat{y}$ is the predicted label calculated using equation~\ref{eq:predict}, in which $m$ is the number of dimensions of the output capsule $\textbf{v}_j$.

\begin{equation}
\label{eq:predict}
 \hat{y} = \frac{1}{m} \sum_i softmax \left({\begin{bmatrix} \textbf{v}_1^\intercal \\ \textbf{v}_2^\intercal \end{bmatrix}}_{:,i}\right)
\end{equation}

The use of equation~\ref{eq:predict} instead of simply using the length of the output capsules~\cite{sabour2017dynamic} promotes separation between the two output capsules on each dimension.

\section{EVALUATION}
\label{sec:eval}
To evaluate the advantage of using random noise, we tested the proposed method with and without using random noise (Capsule-Forensics-Noise and Capsule-Forensics, respectively). The random noise was generated from a normal distribution $N(0, 0.01)$ and was used in the training phase only. Two iterations ($r=2$) were used in the dynamic routing algorithm. We used the half total error rate (HTER) $\big(\frac{FRR + FAR}{2}\big)$ and accuracy $\big( \frac{TP + TN}{TP + TN + FP + FN}\big)$ as metrics.

\subsection{Replay Attack Detection}
To determine the ability of the proposed method to detect replay attacks, we compared its performance with that of eight state-of-the-art detection methods on the well-known Idiap REPLAY-ATTACK dataset~\cite{chingovska2012effectiveness}. As shown in Table~\ref{tab:present}, the proposed method with random noise (Capsule-Forensics-Noise), as well as our previous method~\cite{nguyen2018modular}, had an HTER of zero.

\begin{table}[h!]
\centering
\caption{Half total error rate of state-of-the-art detection methods on REPLAY-ATTACK dataset~\cite{chingovska2012effectiveness}.}
\label{tab:present}
\ninept
\begin{tabular}{l|c}
\multicolumn{1}{c|}{\textbf{Method}} & \textbf{HTER (\%)} \\ \hline
Chigovska et al.~\cite{chingovska2012effectiveness} & 17.17 \\
Pereira et al.~\cite{de2013can} & 08.51 \\
Kim et al.~\cite{kim2015face} & 12.50 \\
Yang et al.~\cite{yang2014learn} & 02.30 \\
Menotti et al.~\cite{menotti2015deep} & 00.75 \\
Alotabib et al.~\cite{alotaibi2017deep} & 10.00 \\
Ito et al.~\cite{ito2017recent} & 00.43 \\
Nguyen et al.~\cite{nguyen2018modular} & \textbf{00.00} \\
Capsule-Forensics & 00.28 \\
Capsule-Forensics-Noise & \textbf{00.00}
\end{tabular}
\end{table}

\subsection{Face Swapping Detection}
We determined the ability of our proposed method to detect face swapping using a deepfake technique on the deepfake dataset proposed by Afchar et al.~\cite{afchar2018mesonet} at both the frame and video levels. As shown in Tables~\ref{tab:deepfake_img} and~\ref{tab:deepfake_vid}, our proposed method with random noise (Capsule-Forensics-Noise) had the highest accuracy in both cases.

\begin{table}[th!]
\centering
\caption{Accuracy of face swapping detection at frame level on deepfake dataset~\cite{afchar2018mesonet}.}
\label{tab:deepfake_img}
\ninept
\begin{tabular}{l|c}
\multicolumn{1}{c|}{\textbf{Method}} & \textbf{Accuracy (\%)} \\ \hline
Meso-4~\cite{afchar2018mesonet} & 89.10 \\
MesoInception-4~\cite{afchar2018mesonet} & 91.70 \\
Nguyen et al.~\cite{nguyen2018modular} & 92.36 \\
Capsule-Forensics & 94.47 \\
Capsule-Forensics-Noise & \textbf{95.93}
\end{tabular}
\end{table}

\begin{table}[th!]
\centering
\caption{Accuracy of face swapping detection at video level on deepfake dataset~\cite{afchar2018mesonet}.}
\label{tab:deepfake_vid}
\ninept
\begin{tabular}{l|c}
\multicolumn{1}{c|}{\textbf{Method}} & \textbf{Accuracy (\%)} \\ \hline
Meso-4~\cite{afchar2018mesonet} & 96.90 \\
MesoInception-4~\cite{afchar2018mesonet} & 98.40 \\
Capsule-Forensics & 97.69 \\
Capsule-Forensics-Noise & \textbf{99.23}
\end{tabular}
\end{table}

\subsection{Facial Reenactment Detection}
We determined the ability of our proposed method to detect facial reenactment on the FaceForensics dataset~\cite{rossler2018faceforensics}, which was created using the Face2Face method~\cite{thies2016face2face}. We strictly followed the authors' guidelines for processing the data. As shown in Table~\ref{tab:faceforensics_img}, on average, the proposed method (with and without noise) had performance comparable to that of the best-performing state-of-the-art methods.

We also tested our method at the video level and compared its performance with that of Afchar et al.'s MesoNet facial video forgery detection network~\cite{afchar2018mesonet}. For our method, we used only the first ten frames instead of the entire video. As shown in Table~\ref{tab:faceforensics_vid}, our method outperformed Afchar et al.'s network.

\subsection{Fully Computer-Generated Image Detection}
Finally, we compared the performance of our proposed method with that of state-of-the-art methods on computer-generated images (CGIs) and photographic images (PIs) on the dataset proposed by Rahmouni et al.~\cite{rahmouni2017distinguishing}. Once again, as shown in Table~\ref{tab:cgi_pi}, our method had the best performance and had perfect accuracy on full-size test images.

\begin{table}[th!]
\centering
\caption{Accuracy of state-of-the-art facial reenactment detection methods at frame level on FaceForensics dataset~\cite{rossler2018faceforensics} with three levels of compression: no compression, easy compression (23), and strong compression (40).}
\label{tab:faceforensics_img}
\ninept
\begin{tabular}{l|c|c|c}
\multicolumn{1}{c|}{\multirow{2}{*}{\textbf{Method}}} & \multicolumn{3}{c}{\textbf{Accuracy (\%)}} \\ \cline{2-4} 
 & \multicolumn{1}{l|}{No-C} & \multicolumn{1}{l|}{Easy-C} & \multicolumn{1}{l}{Hard-C} \\ \hline
Fridrich \& Kodovsky~\cite{fridrich2012rich} & 99.40 & 75.87 & 58.16 \\
Cozzolino et al.~\cite{cozzolino2017recasting} & 99.60 & 79.80 & 55.77 \\
Bayar \& Stamm~\cite{bayar2016deep} & 99.53 & 86.10 & 73.63 \\
Rahmouni et al.~\cite{rahmouni2017distinguishing} & 98.60 & 88.50 & 61.50 \\
Raghavendra et al.~\cite{raghavendra2017transferable} & 97.70 & 93.50 & 82.13 \\
Zhou et al.~\cite{zhou2017two} & 99.93 & 96.00 & 86.83 \\
Rossler et al.~\cite{rossler2018faceforensics} & 99.93 & 98.13 & 87.81 \\
Meso-4~\cite{afchar2018mesonet} & 94.60 & 92.40 & 83.20 \\
MesoInception-4~\cite{afchar2018mesonet} & 96.80 & 93.40 & 81.30 \\
Nguyen et al.~\cite{nguyen2018modular} & 98.80 & 96.10 & 76.40 \\
Capsule-Forensics & 99.13 & 97.13 & 81.20 \\
Capsule-Forensics-Noise & 99.37 & 96.50 & 81.00
\end{tabular}
\end{table}

\begin{table}[th!]
\centering
\caption{Comparison with MesoNet network at video level on FaceForensics dataset~\cite{rossler2018faceforensics}.}
\label{tab:faceforensics_vid}
\ninept
\begin{tabular}{l|c|c|c}
\multicolumn{1}{c|}{\multirow{2}{*}{ }} & \multicolumn{3}{c}{\textbf{Accuracy (\%)}} \\ \cline{2-4} 
 & \multicolumn{1}{l|}{No-C} & \multicolumn{1}{l|}{Easy-C} & \multicolumn{1}{l}{Hard-C} \\ \hline
Meso-4~\cite{afchar2018mesonet} & - & 95.30 & - \\
MesoInception-4~\cite{afchar2018mesonet} & - & 95.30 & - \\
Capsule-Forensics & \textbf{99.33} & \textbf{98.00} & 82.00 \\
Capsule-Forensics-Noise & \textbf{99.33} & 96.00 & \textbf{83.33}
\end{tabular}
\end{table}

\begin{table}[th!]
\centering
\caption{Accuracy of state-of-the-art methods on discriminating between CGIs and PIs.}
\label{tab:cgi_pi}
\ninept
\begin{tabular}{l|c|c}
\multicolumn{1}{c|}{\multirow{2}{*}{\textbf{Method}}} & \multicolumn{2}{c}{\textbf{Accuracy}} \\ \cline{2-3} 
\multicolumn{1}{c|}{} & \textbf{Patch} & \textbf{Full Size} \\ \hline
Rahmouni et al.~\cite{rahmouni2017distinguishing} & 89.76 & 99.30 \\
Quan et al.~\cite{quan2018distinguishing} & 94.75 & 99.58 \\
Nguyen et al.~\cite{nguyen2018modular} & 96.55 & 99.86 \\
Capsule-Forensics & 96.75 & 99.72 \\
Capsule-Forensics-Noise & \textbf{97.00} & \textbf{100.00}
\end{tabular}
\end{table}

\section{CONCLUSION}
\label{sec:conclusion}
Our comprehensive experiments demonstrated the feasibility of building a general detection method that is effective for a wide range of forged image and video attacks. They also demonstrated that capsule networks can be used in domains other than computer vision. The proposed use of random noise in the training phase proved beneficial in most cases. Future work will mainly focus on evaluating the ability of the proposed method to resist adversarial machine attacks, especially on the proposed random noise at test time, and enhancing its ability. It will also focus on making the proposed method robust against mixed attacks and on raising this critical issue in the research community.

\section{ACKNOWLEDGMENTS}
This work was supported by JSPS KAKENHI Grant Numbers (16H06302, 17H04687, 18H04120, 18H04112, 18KT0051) and by JST CREST Grant Number JPMJCR18A6, Japan.
\section{PREFERENCES}
\bibliographystyle{IEEEbib}
\bibliography{refs}

\end{document}